%% file: main.tex
\documentclass[11pt]{article}

\usepackage[final]{acl}

\usepackage{times}
\usepackage{latexsym}
\usepackage[T1]{fontenc}
\usepackage[utf8]{inputenc}
\usepackage{microtype}
\usepackage{inconsolata}
\usepackage{graphicx}
\usepackage{amsmath,amssymb}
\usepackage{booktabs}
\usepackage{multirow}
\usepackage{xcolor}

\newcommand{\grpo}{\textsc{GRPO}}
\newcommand{\lora}{\textsc{LoRA}}

\title{When Gradient Importance Lies:\\Adaptive {LoRA} Rank Allocation Fails Under {GRPO}}

\author{Yash Ganpat Sawant \\
  Independent AI Researcher \\
  \texttt{sawantyash13@gmail.com} \\}

\hypersetup{
  pdftitle={When Gradient Importance Lies: Adaptive LoRA Rank Allocation Fails Under GRPO},
  pdfauthor={Yash Ganpat Sawant}
}

\begin{document}
\maketitle

\begin{abstract}
Adaptive rank allocation for \lora{}---allocating more parameters to important layers and fewer to unimportant ones---consistently improves efficiency under supervised fine-tuning (SFT).
We test whether this success transfers to reinforcement learning, specifically Group Relative Policy Optimization (\grpo{}).
Using gradient-magnitude profiling on Qwen~2.5 1.5B with GSM8K, we find that, \emph{in our setting}, it does not: proportional rank allocation \emph{degrades} accuracy by 4.5 points compared to uniform allocation (70.0\% vs.\ 74.5\%), despite using identical parameter budgets.
We identify two mechanisms behind this failure.
First, the gradient landscape under \grpo{} is fundamentally flatter than under SFT---the max-to-min layer importance ratio is only $2.17\times$, whereas the layer concentration reported by \citet{shi2024ila} for SFT (top 30\% of layers carrying $>$80\% of the gradient signal) implies a max/min ratio well above $10\times$.
All layers carry meaningful gradient signal; none are truly idle.
Second, we observe a \emph{gradient amplification effect}: non-uniform allocation widens the importance spread from $2.17\times$ to $3.00\times$, creating a positive feedback loop where high-rank layers absorb more gradient while low-rank layers are progressively silenced.
A random-allocation control yields the same amplification ($r=0.972$ correlation between assigned rank and resulting gradient share), indicating that rank causally determines gradient importance rather than the reverse.
The negative result is single-seed and single-task; we present it as preliminary evidence that gradient importance does not predict capacity requirements under RL, and that na\"ive transfer of SFT-era rank allocation strategies to alignment training should be evaluated cautiously.
\end{abstract}

\section{Introduction}

Parameter-efficient fine-tuning via Low-Rank Adaptation \citep{hu2022lora} is the standard approach for adapting large language models.
\lora{} decomposes weight updates into low-rank matrices $\Delta W = BA$ with uniform rank $r$ across all layers.
A line of follow-up work has challenged this uniform assumption.
AdaLoRA \citep{zhang2023adalora} dynamically prunes singular values during training based on importance scores.
GoRA \citep{gora2025} allocates rank proportionally to gradient–weight products at initialization.
Aletheia \citep{aletheia2026} selects layers via lightweight gradient probes.
ILA \citep{shi2024ila} shows that only 10--30\% of layers carry the bulk of the gradient signal for alignment under SFT, with the top 30\% carrying $>$80\% of total magnitude.
These methods report meaningful efficiency gains, supporting the picture that \textbf{gradient importance correlates with capacity requirements under supervised objectives}.

A natural question follows: does this correlation hold under reinforcement learning?
RL-based alignment methods like \grpo{} \citep{shao2024deepseekmath} optimize an advantage-weighted policy gradient with sparse, often binary reward, rather than a dense per-token cross-entropy loss.
Theoretical work suggests RL gradients concentrate differently than SFT gradients \citep{shallow2026}, motivating the hypothesis that rank allocation strategies should differ.

\textbf{We report a negative result.}
In our setting---Qwen~2.5 1.5B fine-tuned on GSM8K with \grpo{}---applying the same gradient-magnitude profiling and proportional allocation recipe that helps SFT \emph{hurts} \grpo{} accuracy by 4.5 points relative to uniform allocation at an identical parameter budget.
A random control degrades further, but uniform allocation outperforms both.
The result is preliminary (one model, one dataset, single seed, $n=200$ eval), but the mechanism we expose is robust across our four configurations and we believe it is worth surfacing.

We make three observations:
\begin{enumerate}
    \item \textbf{Flat gradient landscape.} Under \grpo{} on this task, layer importance is distributed far more uniformly than what \citet{shi2024ila} report for SFT (max/min ratio $2.17\times$ vs.\ an implied $>$10$\times$). All layers carry load.
    \item \textbf{Gradient amplification.} Non-uniform allocation widens the importance spread from $2.17\times$ to $3.00\times$. High-rank layers absorb more gradient; low-rank layers are silenced. Rank thus causally determines the observed gradient distribution.
    \item \textbf{Generalization gap.} Models train equally well across allocations (Fig.~\ref{fig:reward_curves}), but the proportional and random configurations fail to transfer to the held-out test set. The damage from rank reallocation appears only at evaluation time.
\end{enumerate}

We frame these findings as a case study in why SFT-era efficiency tricks should not be transferred to RL without empirical validation.

\section{Method}

\subsection{GRPO with LoRA}

\grpo{} generates $K$ completions per prompt, computes rewards, and normalizes advantages within the group:
\begin{equation}
    \hat{A}_i = \frac{r_i - \mu_{\text{group}}}{\sigma_{\text{group}}}
\end{equation}
Combined with \lora{}, gradients flow only through the adapter parameters $B^{(l)}, A^{(l)}$ at each layer $l$.

\subsection{Reward Sensitivity Profiling}

We define the reward sensitivity score for layer $l$ as the mean over $T$ training steps of the summed adapter gradient norm:
\begin{equation}
    S(l) = \frac{1}{T} \sum_{t=1}^{T} \sum_{m \in \mathcal{M}^{(l)}} \left\| \nabla_{\theta_m^{(l)}} \mathcal{L}_t \right\|_2 \label{eq:profile}
\end{equation}
where $\mathcal{M}^{(l)}$ is the set of target modules at layer $l$ (q/k/v/o/up/down/gate projections). Equation~\ref{eq:profile} aggregates over modules within a layer; the per-module decomposition we use later for the attention vs.\ FFN analysis (\S\ref{sec:why}) is computed separately and reported in Appendix~\ref{app:module}.

\subsection{Rank Allocation}

Given a total rank budget $R_{\text{total}} = L \times r_{\text{uniform}}$, we first normalize the sensitivity score, $\hat{s}(l) = S(l) / \sum_{l'} S(l')$, and then allocate per-layer rank
\begin{equation}
    r^{(l)} = \text{clip}\big(\text{round}(\hat{s}(l) \cdot R_{\text{total}}), r_{\min}, r_{\max}\big)
\end{equation}
with $r_{\min}=4$, $r_{\max}=64$, rounded to multiples of 4.
As a control we also evaluate \emph{random} allocation: a permutation of the proportional rank vector across layers, removing any link to $S(l)$ while preserving the rank distribution and budget.
A \emph{reduced 70\%} variant scales the proportional vector to 70\% of the uniform budget ($r\in[12,31]$, $25.8$M trainable parameters vs.\ $36.9$M for the same-budget conditions) to probe whether smaller models can match uniform performance.

\section{Experiments}

\subsection{Setup}
\label{sec:setup}

\textbf{Model:} Qwen/Qwen2.5-1.5B-Instruct (28 transformer layers).
\textbf{Dataset:} GSM8K \citep{cobbe2021gsm8k} with structured XML output (\texttt{<think>}/\texttt{<answer>}).
\textbf{Rewards:} Format compliance (1.0 for correctly closed tags) and answer correctness (1.0 for the correct numerical answer).
\textbf{LoRA:} Applied to all seven projection modules per layer (196 adapters total). Uniform baseline: $r=32$, giving 36.9M trainable parameters.
Across every configuration we scale the \lora{} $\alpha$ per layer as $\alpha^{(l)}=2\,r^{(l)}$, so the effective update scale $\alpha/r=2$ is held constant regardless of the rank assigned to a layer; the accuracy gaps we report are therefore not an artifact of varying per-layer effective learning rates.
\textbf{GRPO:} $K=4$ generations per prompt, $\beta=0.05$ KL coefficient, 1000 training steps, learning rate $10^{-5}$, single seed.
We run rollout generation in-process with vLLM (\emph{colocate} mode: the policy used for sampling shares GPU memory with the trainer, avoiding network transfer).
\textbf{Eval:} 200 held-out GSM8K test problems; greedy decoding; Wilson 95\% intervals.

\subsection{Gradient Landscape Under GRPO}

\begin{figure}[t]
    \centering
    \includegraphics[width=0.95\columnwidth]{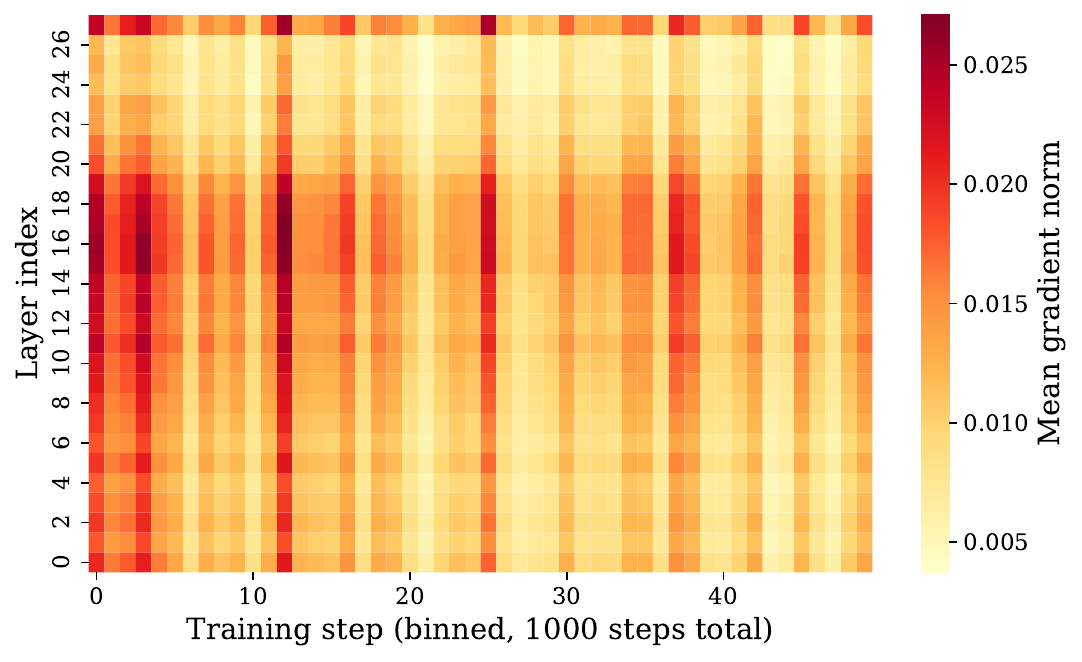}
    \caption{Per-layer adapter gradient magnitude during \grpo{} training (1000 steps, 28 layers, uniform $r=32$). The distribution is notably flat relative to SFT-era reports.}
    \label{fig:heatmap}
\end{figure}

Figure~\ref{fig:heatmap} shows the reward sensitivity map under uniform \lora{}.
We note four properties:
\begin{itemize}
    \item \textbf{Flat distribution.} Max/min importance ratio is $2.17\times$ (Layer~15 hottest at 4.68\%; Layer~26 coldest at 2.15\%). Even the coldest layer carries 46\% of the hottest layer's gradient signal.
    \item \textbf{Middle-layer concentration.} Layers 9--18 carry 43.0\% of the gradient, but early (29.8\%) and late (27.2\%) layers also remain meaningful---unlike the SFT picture, where ILA \citep{shi2024ila} reports $>$80\% concentration in the top 30\% of layers.
    \item \textbf{Temporal stability.} The correlation between the per-layer importance profile in the first 250 steps and the last 250 steps is 0.962, indicating stable structural patterns rather than transient noise.
    \item \textbf{Module composition.} Attention modules (q/k/v/o) account for 52.2\% of the gradient signal, FFN modules (up/down/gate) for 47.8\%. The \texttt{up\_proj} module is the single most reward-sensitive (21.7\%; per-module breakdown in Appendix~\ref{app:module}).
\end{itemize}

\subsection{Rank Allocation Results}

\begin{table}[t]
    \centering
    \small
    \begin{tabular}{lccc}
        \toprule
        \textbf{Strategy} & \textbf{Params} & \textbf{Acc.\ (\%)} & \textbf{95\% CI} \\
        \midrule
        Base (no LoRA) & 0 & 66.0 & $\pm$6.6 \\
        \midrule
        Uniform ($r{=}32$) & 36.9M & \textbf{74.5} & $\pm$6.0 \\
        Proportional ($r{=}20\text{--}40$) & 36.9M & 70.0 & $\pm$6.4 \\
        Random ($r{=}16\text{--}48$) & 36.9M & 67.5 & $\pm$6.5 \\
        \midrule
        Reduced 70\% ($r{=}12\text{--}31$) & 25.8M & 65.0 & $\pm$6.6 \\
        \bottomrule
    \end{tabular}
    \caption{GSM8K accuracy under different rank allocations ($n{=}200$ test samples; single seed). All ``same budget'' methods use total rank 896 ($28 \times 32$). Confidence intervals are Wilson score intervals.}
    \label{tab:results}
\end{table}

Table~\ref{tab:results} shows our main result: \textbf{uniform allocation outperforms every non-uniform variant we tested}, including gradient-aware proportional allocation, at the same parameter budget.
Proportional allocation scores 4.5 points below uniform.
Random allocation scores even lower (67.5\%).
Reduced-budget allocation falls below the untrained base model.
The confidence intervals overlap, so we treat the magnitudes as suggestive rather than statistically decisive (see Limitations).

\begin{figure}[t]
    \centering
    \includegraphics[width=0.95\columnwidth]{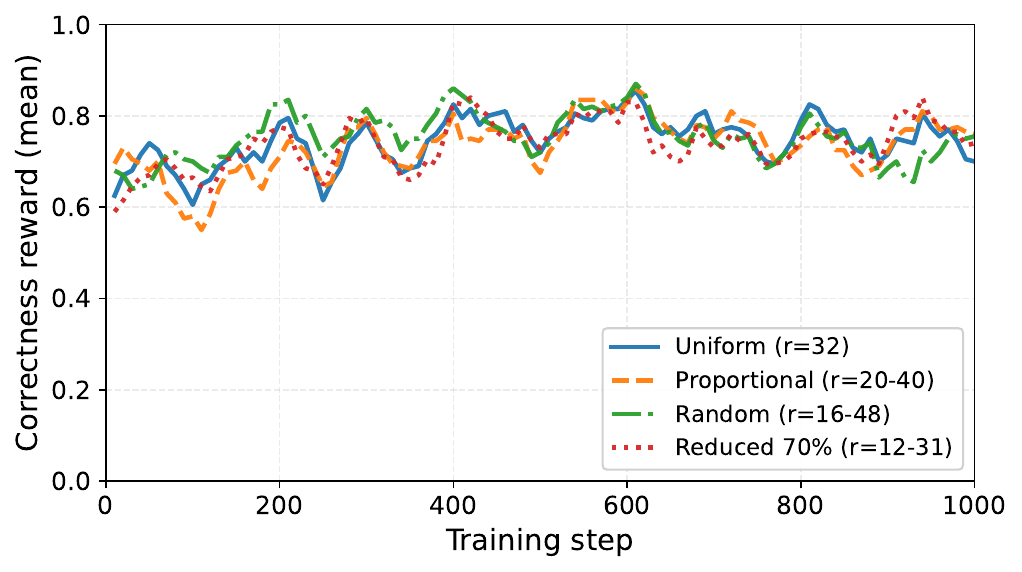}
    \caption{Training reward curves are nearly indistinguishable across all four configurations (Uniform, Proportional, Random, Reduced~70\%); the evaluation gap in Table~\ref{tab:results} emerges only on the held-out set.}
    \label{fig:reward_curves}
\end{figure}

Two observations stand out.
First, \textbf{gradient-aware proportional allocation outperforms random allocation} (70.0\% vs.\ 67.5\%), so the importance signal is not noise---it identifies layers that matter.
Yet both lose to uniform, so the signal, while directionally correct, is not actionable as a rank allocator under \grpo{}.
Second, training reward curves are nearly identical across all four configurations (Fig.~\ref{fig:reward_curves}): correctness reward reaches $0.74$--$0.77$ for every method by step 1000.
The performance gap appears \emph{only at evaluation}, suggesting that rank reallocation harms generalization rather than training dynamics.

\subsection{The Gradient Amplification Effect}

\begin{figure}[t]
    \centering
    \includegraphics[width=0.95\columnwidth]{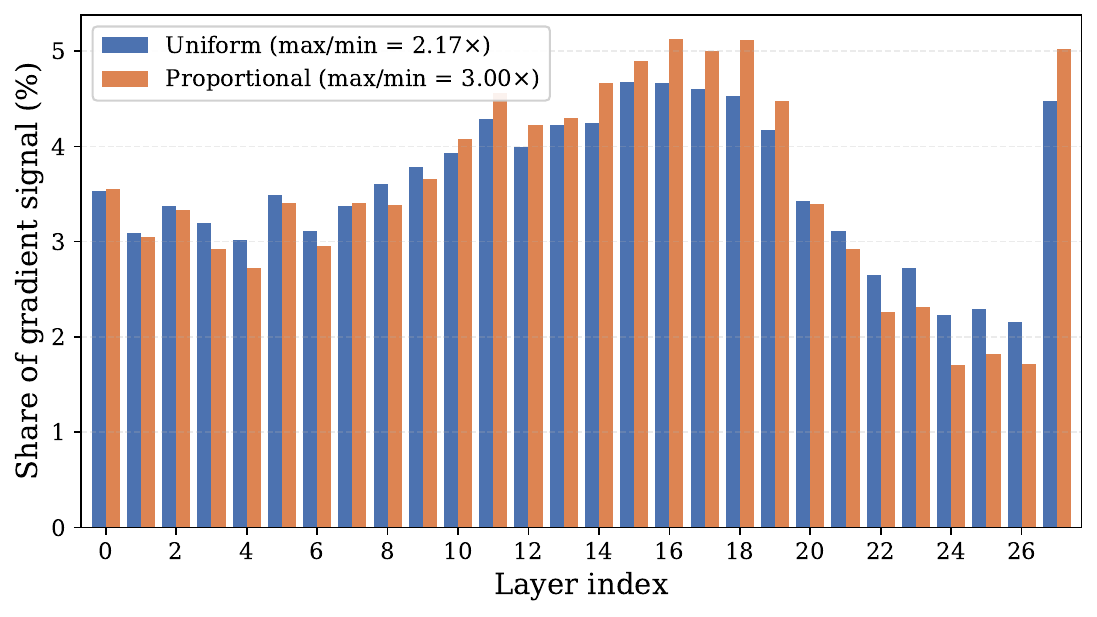}
    \caption{Normalized layer importance under uniform vs.\ proportional allocation. Non-uniform allocation amplifies the original importance spread from $2.17\times$ to $3.00\times$.}
    \label{fig:amplification}
\end{figure}

We profile gradients during \emph{all} training runs, not just the uniform baseline.
Figure~\ref{fig:amplification} reveals a striking effect: the gradient importance spread \emph{widens} under non-uniform allocation.

\begin{itemize}
    \item Uniform: max/min ratio $2.17\times$.
    \item Proportional (same budget): $3.00\times$ ($+38\%$).
    \item Reduced 70\%: $3.57\times$ ($+64\%$).
\end{itemize}

Layers given higher rank (e.g.\ Layer~18: $r=40$) see their gradient share \emph{increase} from 4.53\% to 5.12\%.
Conversely, layers given lower rank (e.g.\ Layer~24: $r=20$) see their share \emph{decrease} from 2.23\% to 1.71\%.
The allocation creates a positive feedback loop: more capacity $\rightarrow$ more gradient $\rightarrow$ appears even more ``important''.

Critically, this effect is \textbf{causal, not correlative}.
The random allocation experiment---where ranks are deliberately decoupled from gradient importance---shows the same amplification.
The correlation between assigned rank and resulting gradient share is $0.972$ for random and $0.946$ for proportional.
Under random allocation, Layer~1 (normally 3.09\% of gradient) receives $r=48$ and jumps to 4.21\%; Layer~10 (normally 3.93\%) receives $r=16$ and drops to 2.85\%.
\textbf{Rank determines gradient importance, not the other way around.}

This has a methodological consequence: gradient profiling under one allocation cannot reliably inform a different allocation. The ``profile then retrain'' paradigm common in SFT-era adaptive-rank methods is unstable for \grpo{}, at least at this scale.
One natural concern is whether the effect is trivially explained by more parameters yielding larger aggregate gradient norms. Equation~\ref{eq:profile} aggregates gradient norms within a layer, so larger layers can mechanically produce larger $S(l)$. However, the same amplification persists when restricted to individual modules at identical dimensions (e.g.\ \texttt{q\_proj} at $1536\times 1536$ across all layers; Appendix~\ref{app:module}), so it is not purely a parameter-counting artifact.

\subsection{Why SFT Methods May Fail Under RL}
\label{sec:why}

We sketch an interpretation. Under SFT the loss is per-token cross-entropy and a small number of layers dominate the gradient landscape (\citealp{shi2024ila}: $>$80\% in the top 30\%), so there are genuinely idle layers whose capacity can be redistributed. Under \grpo{}, the loss is an advantage-weighted policy gradient with sparse, binary reward; the top 30\% of layers in our profile carry only 35.7\% of the signal, and even the coldest layers contribute 2.15\%. Reducing their capacity---even modestly, from $r=32$ to $r=20$---damages evaluation accuracy while leaving training reward unchanged. We conjecture that ``cold'' layers in this setting handle structural functions (output formatting, numerical precision, coherence) that are already well-handled by the pretrained model and become bottlenecks only when capacity is reduced. The near-even attention/FFN split (52.2\%/47.8\%) is consistent with this story, unlike SFT where FFN layers in the upper half dominate \citep{zhang2023adalora}.

\section{Related Work}

\textbf{Adaptive rank for SFT.}
AdaLoRA \citep{zhang2023adalora} prunes singular values during training.
GoRA \citep{gora2025} uses gradient–weight products for initialization-time allocation.
IGU-LoRA \citep{igulora2026} applies integrated gradients with uncertainty-aware scoring.
Aletheia \citep{aletheia2026} selects layers via gradient probes.
All of these operate exclusively under supervised objectives.

\textbf{Layer importance in alignment.}
ILA \citep{shi2024ila} learns binary layer masks and shows that 10--30\% of layers suffice for SFT alignment.
\citet{shallow2026} argue that RLHF gradients concentrate at specific positions.
Our work extends this line by showing that even when RL gradients are non-uniform, their concentration is too mild to support useful rank reallocation.

\textbf{GRPO and LoRA.}
DeepSeekMath \citep{shao2024deepseekmath} introduced \grpo{} for mathematical reasoning.
To our knowledge, no prior work has investigated adaptive rank allocation specifically under RL-based alignment.

\section{Limitations}

The result is \emph{preliminary} and the following caveats matter. The experiments use one model (Qwen~2.5 1.5B), one dataset (GSM8K), one RL algorithm (\grpo{}), and single-seed runs with $n=200$ test problems. The Wilson intervals in Table~\ref{tab:results} overlap, so the 4.5-point gap is suggestive rather than statistically decisive. We compare to SFT gradient distributions only via prior work \citep{shi2024ila,zhang2023adalora}; a matched SFT baseline on the same model and data would sharpen the contrast. We hold \lora{} scaling constant across allocations ($\alpha/r=2$ per layer), but have not isolated the contribution of adapter initialization or optimizer state from the rank itself. Finally, our cold-layer hypothesis is untested; per-reward gradient decomposition (correctness vs.\ format) could partially verify it. We encourage replication at larger scale.

\section{Lessons and Conclusion}

In our setting, the gradient-based rank allocation that works under SFT does not transfer to RL alignment via \grpo{}: proportional allocation degrades accuracy by 4.5 points at an identical parameter budget, random allocation degrades further, and a random control shows that rank itself causally shapes the resulting gradient distribution. The mechanism---a flat gradient landscape combined with a positive-feedback amplification effect---suggests three practitioner-facing lessons. (i) Gradient importance is a useful diagnostic but not an allocator under \grpo{}: the signal ranks layers correctly (proportional beats random) but does not justify deviating from uniform at a fixed budget. (ii) Profile-then-retrain is unstable under RL, because the profile changes with the allocation ($r=0.97$ between assigned rank and post-hoc importance). (iii) Methods that exploit concentrated gradient structure under SFT cannot be assumed to work under RL, where the structure we observe is much flatter. Future work should test whether dynamic, training-time rank adaptation can break the feedback loop, and whether the effect persists at larger scale, on other tasks, and under other RL objectives such as PPO and DPO.

\bibliography{references}

\appendix
\section{Per-Module Gradient Importance}
\label{app:module}

Table~\ref{tab:modules} reports the average per-module gradient magnitude across all 28 layers under the uniform $r=32$ baseline.
Values are aggregated from the $B$ adapter matrix at each layer (the $A$ matrix shows the same qualitative ordering at smaller magnitude) and averaged over the final 200 steps of \grpo{} training.
The split between attention (q/k/v/o) and FFN (up/down/gate) modules is near-even (52.2\% vs.\ 47.8\%), with \texttt{up\_proj} the single most reward-sensitive module.

\input{module_table.tex}

\end{document}

%% file: module_table.tex
\begin{table}[t]
  \centering
  \small
  \begin{tabular}{llcl}
    \toprule
    \textbf{Module} & \textbf{Family} & \textbf{Share (\%)} & \textbf{Top-3 layers} \\
    \midrule
    \texttt{up\_proj} & FFN & 21.7 & L27, L3, L18 \\
    \texttt{gate\_proj} & FFN & 17.1 & L27, L18, L17 \\
    \texttt{v\_proj} & Attn & 15.6 & L0, L11, L15 \\
    \texttt{o\_proj} & Attn & 14.9 & L0, L15, L16 \\
    \texttt{k\_proj} & Attn & 12.4 & L18, L17, L19 \\
    \texttt{q\_proj} & Attn & 9.3 & L16, L14, L9 \\
    \texttt{down\_proj} & FFN & 8.9 & L1, L27, L2 \\
    \midrule
    \multicolumn{2}{l}{Attention total} & 52.2 & --- \\
    \multicolumn{2}{l}{FFN total} & 47.8 & --- \\
    \bottomrule
  \end{tabular}
  \caption{Per-module share of total adapter gradient magnitude under uniform $r=32$ \grpo{} training on GSM8K (mean over final 200 steps; $B$ matrices). The attention/FFN split is near-even (52.2\%/47.8\%), and \texttt{up\_proj} is the single most reward-sensitive module.}
  \label{tab:modules}
\end{table}